\documentclass[runningheads]{llncs}
\usepackage{graphicx}
\usepackage{booktabs}
\usepackage{caption}
\usepackage{rotating}
\usepackage{multirow}
\usepackage{booktabs}
\usepackage{adjustbox}
\usepackage{pifont}
\usepackage{xcolor}
\usepackage{booktabs}
\usepackage{xcolor}
\usepackage{amsmath}
\usepackage{amssymb}
\usepackage{multirow}
\begin{document}
\vspace{-5cm}   
\title{Semi-KAN: KAN Provides an Effective Representation for Semi-Supervised Learning in Medical Image Segmentation}
\titlerunning{Semi-KAN}
\author{Zanting Ye, Xiaolong Niu, Xuanbin Wu, Wenxiang Yi, Yuan Chang, Lijun Lu*}  
\authorrunning{Ye et al.}
\institute{School of Biomedical Engineering, Southern Medical University \\
    \email{ljlubme@gmail.com}}

%
%
\maketitle           
\begin{abstract}
Deep learning-based medical image segmentation has shown remarkable success; however, it typically requires extensive pixel-level annotations, which are both expensive and time-intensive. Semi-supervised medical image segmentation (SSMIS) offers a viable alternative, driven by advancements in Convolutional Neural Networks (CNNs) and Vision Transformers (ViTs). Nevertheless, these networks often rely on single fixed activation functions and linear modeling patterns, limiting their ability to effectively learn robust representations. Given the limited availability of labeled data (e.g., 5\% or 10\%), achieving robust representation learning becomes crucial. Inspired by Kolmogorov-Arnold Networks (KANs), we propose Semi-KAN, which leverages the untapped potential of KANs to enhance backbone architectures for representation learning in SSMIS. Our findings indicate that: (1) compared to networks with fixed activation functions, KANs exhibit superior representation learning capabilities with fewer parameters, and (2) KANs excel in high-semantic feature spaces. Building on these insights, we integrate KANs into tokenized intermediate representations, applying them selectively at the encoder's bottleneck and the decoder's top layers within a U-Net pipeline to extract high-level semantic features.Although learnable activation functions improve feature expansion, they introduce significant computational overhead with only marginal performance gains. To mitigate this, we reduce the feature dimensions and employ horizontal scaling to capture multiple pattern representations. Furthermore, we design a multi-branch U-Net architecture with uncertainty estimation to effectively learn diverse pattern representations. Extensive experiments on four public datasets demonstrate that Semi-KAN surpasses baseline networks, utilizing fewer KAN layers and lower computational cost, thereby underscoring the potential of KANs as a promising approach for SSMIS.
\end{abstract}

\begin{figure*}[t]
\centering
\includegraphics[width=1\textwidth]{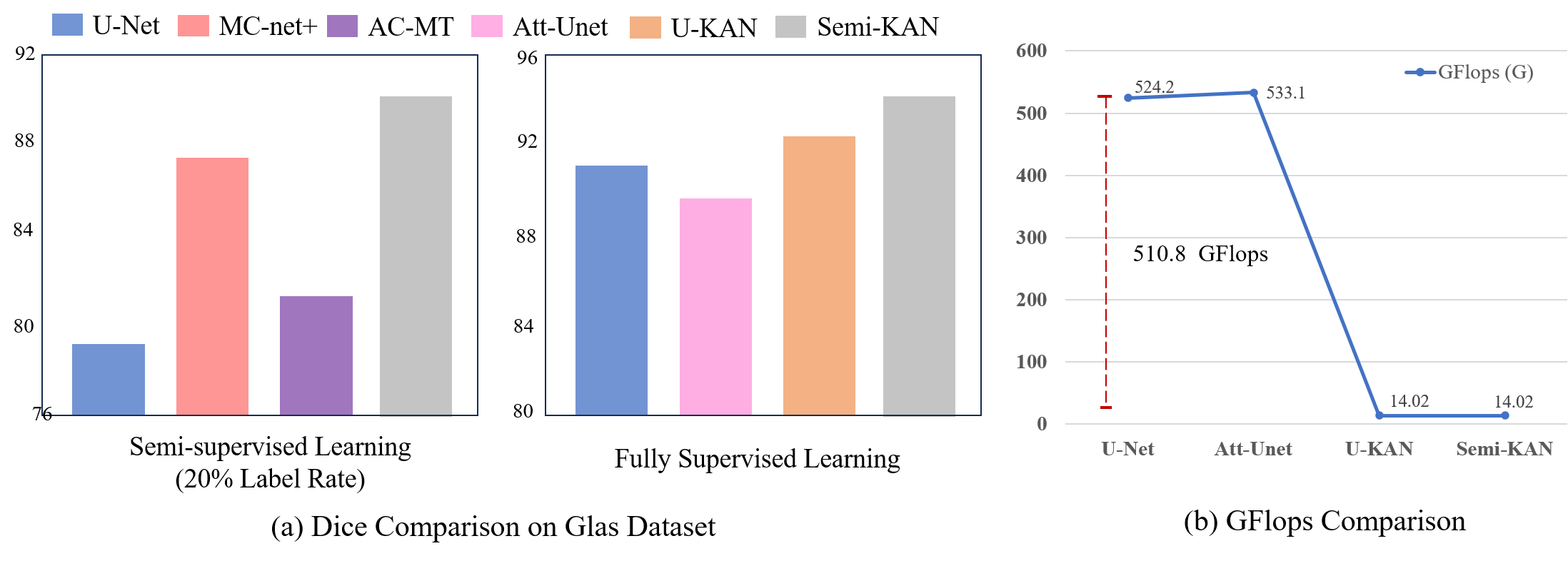} 
\caption{Performance and efficiency comparison of U-Net, Att-Unet, U-KAN, and Semi-KAN on the GLAS dataset. GFLOPs are measured during model inference with an input size of 256 $\times$ 256.}
\label{fig1}
\end{figure*}

\section{Introduction}
Deep learning-based medical image segmentation is a fundamental and critical step in computer-aided diagnosis systems, facilitating the precise identification and quantification of anatomical structures and pathological regions \cite{zhou2023transformer,ma2024segment,zhou2023unified}. Recent advancements in Convolutional Neural Networks (CNNs), Vision Transformers (ViTs), and Mamba have enabled state-of-the-art performance in medical image segmentation across a wide range of clinical applications \cite{wang2024mln,mazurowski2023segment,gu2023mamba}.

Despite these advancements, challenges remain due to the limited availability of annotations. Acquiring such data is labor-intensive and time-consuming, often requiring expert manual pixel-level annotations. As an alternative, self-supervised learning \cite{krishnan2022self,wang2021dense,zhang2023multi} and semi-supervised learning \cite{qiu2023federated,wang2023dual,chen2022semi} have emerged as more efficient approaches. Self-supervised learning, which utilizes only unannotated data, reduces the reliance on manual annotations \cite{zhai2019s4l}. However, while effective for natural images, self-supervised learning faces challenges in medical imaging due to the complex and ambiguous semantic features. Semi-supervised learning leverages a small number of labeled samples to generalize semantic features and adjusts classification boundaries using a large pool of unlabeled data. This approach achieves performance comparable to fully supervised learning and holds significant clinical potential.

Current semi-supervised methods predominantly rely on ViTs and CNNs, employing single fixed activation functions and fully connected feedforward neural networks for feature extraction. However, these methods face fundamental limitations due to suboptimal kernel designs in components such as convolutional layers, transformers, and multi-layer perceptrons (MLPs). These architectures are constrained to linear pattern modeling and channel relationships within the latent space, resulting in limited feature representation capacity.

Recently, Kolmogorov-Arnold Networks (KANs) have incorporated stacks of non-linear, learnable activation functions based on the Kolmogorov-Arnold representation theorem, demonstrating significant potential for enhancing representation capabilities \cite{liu2024kan}. Unlike MLPs, which employ fixed activation functions at each node ("neuron"), KANs introduce learnable activation functions on edges ("weights"). By integrating splines into their architecture, KANs offer a powerful alternative to MLPs, enhancing non-linear representation learning. As illustrated in Fig.~\ref{fig1}, we compare mainstream CNN- and ViT-based medical image segmentation methods with KAN-based approaches in both semi-supervised and fully supervised settings. KANs exhibit superior segmentation performance, underscoring their advantages in semantic feature representation. While learnable activation functions enhance non-linear representation, they also introduce significant computational overhead during training. Therefore, exploring strategies to leverage the advantages of KANs while minimizing computational costs is crucial. Additionally, it remains essential to investigate how the interpretability benefits of KANs can be effectively realized in deep learning networks.

In this work, we explore the application of KANs in SSMIS and propose Semi-KAN. In Semi-KAN, convolution blocks are used to extract local features, while KANs are selectively applied at the bottom of the encoder and the top of the decoder—regions corresponding to high-semantic feature spaces—to capture high-level semantic features. To balance model performance and computational cost, we adopt a multi-mode feature learning strategy and reduce the number of feature layers in single-mode learning. Specifically, we design a shared-encoder, multi-decoder U-Net pipeline with uncertainty estimation-based consistency loss. The multi-decoder architecture facilitates multi-mode extended learning. Semi-KAN combines the local feature extraction advantages of CNNs with the high-level semantic feature extraction capabilities of KANs. As shown in Fig.~\ref{fig1}(b), it improves segmentation performance while maintaining low computational cost. To the best of our knowledge, Semi-KAN represents the first application of KANs to SSMIS, showcasing the superior representation learning capabilities of KANs.

\section{Related Work}
\subsection{Medical Image Segmentation}
Medical image segmentation is a critical component of automated medical image analysis, enabling the extraction of essential quantitative imaging markers to improve diagnosis, personalized treatment planning, and therapy monitoring \cite{sitenko2021assignment}. With the advent of deep learning, segmentation approaches have transitioned from traditional machine learning models to deep learning-based techniques, achieving promising results across various tasks \cite{lecun2015deep,wang2024mln,ma2024segment}. For deep learning-based medical image segmentation, U-Net and its extensions have been widely adopted as baselines for further study, particularly the powerful nnU-Net \cite{isensee2021nnu} and Swin-Unet \cite{cao2022swin}. These extensions focus on enhancing U-Net through effective data augmentation, network design, and loss function optimization. Despite their impressive results, these methods rely on fully supervised learning, which requires large amounts of manually annotated data for training. Manual pixel-level delineation is time-consuming, labor-intensive, and prone to errors and inter-observer variability, making these approaches challenging to deploy in clinical settings \cite{wang2024dual}.

\subsection{Consistency-based Semi-supervised Learning for Medical Image Segmentation}
In the medical imaging domain, annotation scarcity is a significant and inherent challenge. Recently, SSMIS has emerged as a promising approach to address this issue by leveraging vast amounts of unannotated data with supervision from limited labeled data \cite{wang2024towards}. Consistency-based SSMIS has demonstrated strong performance by encouraging high similarity between different predictions for the same input image \cite{luo2021semi,wang2023dual,xu2023ambiguity}. The objective of consistency-based SSMIS is to develop models that are not only accurate in their predictions but also invariant to input or model perturbations, thereby ensuring that the decision boundary traverses the low-density region of the feature space \cite{wu2023exploring}. The key to consistency-based SSMIS lies in discovering effective feature representations under varying perturbations and identifying invariant features. In this work, we introduce Kolmogorov-Arnold Networks (KANs), a network with learnable activation functions, into consistency-based SSMIS. Unlike traditional deep learning networks, which primarily utilize fixed activation functions, KANs employ learnable activation functions to enhance feature representation in unannotated data. We also adopt a multi-mode feature learning framework to further improve the effectiveness of feature representation.

\subsection{Kolmogorov-Arnold Networks}
Multi-layer Perceptrons (MLPs) \cite{cybenko1989approximation}, also known as fully connected feedforward neural networks, serve as foundational building blocks for modern deep learning architectures such as CNNs, ViTs, and Mamba. Recently, Kolmogorov-Arnold Networks (KANs) \cite{liu2024kan} have been proposed as a promising alternative to MLPs. KANs are inspired by the Kolmogorov-Arnold theorem \cite{kolmogorov1957representation}, which states that any continuous function can be represented as a composition of continuous unary functions of finite variables. Unlike MLPs, which rely on fixed activation functions at each node ("neuron"), KANs employ learnable activation functions on edges ("weights"), promoting robust feature representation learning. By integrating splines into their design, KANs provide a powerful alternative to MLPs for non-linear representation learning. KANs have demonstrated effectiveness in approximating high-dimensional, complex functions and robust performance across various applications \cite{li2024u,vaca2024kolmogorov,azam2024suitability}. However, KANs are computationally expensive due to the additional learnable parameters they introduce. In this work, we integrate CNNs with KANs, where CNNs are employed to extract local features and KANs are utilized to capture high-level semantic features. To reduce computational costs and enhance representation learning, we adopt a multi-mode feature learning strategy and minimize the number of feature layers required for single-mode learning.

\section{Main Methodology}
\begin{figure*}[t]
\centering
\includegraphics[width=1\textwidth]{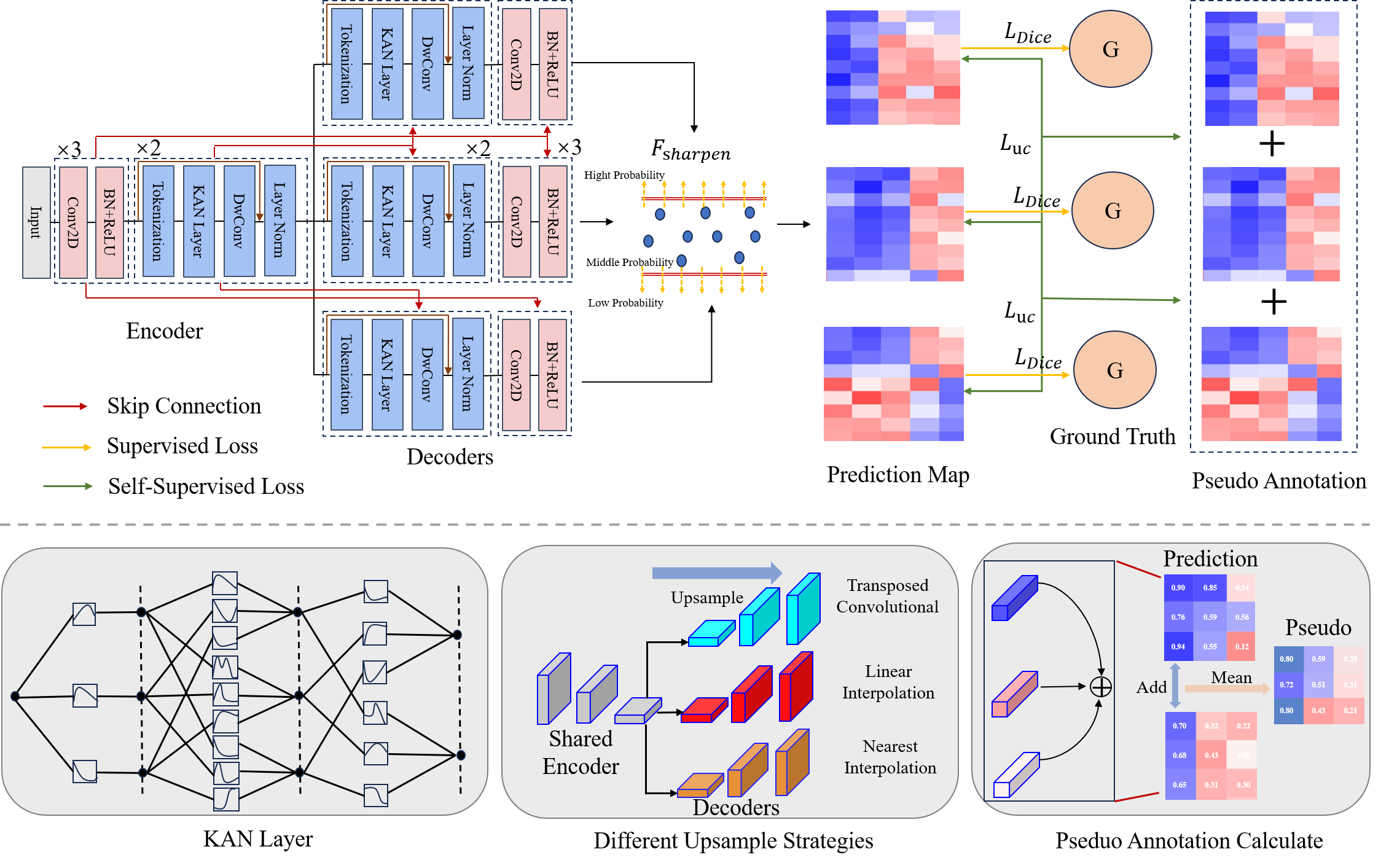} 
\caption{Workflow of the proposed Semi-KAN. Semi-KAN utilizes a shared encoder and a multi-decoder network architecture with KANs. The decoders apply different upsampling strategies as perturbations, and the average of the predicted probability maps from the outputs is used as pseudo-annotations (intermediary supervision annotations).}
\label{fig2}
\end{figure*}

In this section, we present Semi-KAN, a novel KAN-enabled SSMIS approach, as illustrated in Fig.~\ref{fig2}. Semi-KAN is built upon a KAN-based U-Net pipeline network, incorporating dice loss for annotated data and uncertainty estimation-based consistency loss for unannotated data. The following subsections provide a detailed explanation of the implementation and components of the Semi-KAN framework.

\subsection{Semi-KAN Architecture}
An overview of the Semi-KAN architecture is provided in Fig.~\ref{fig2}. Semi-KAN employs a shared encoder and multiple independent decoders, where various upsampling strategies are introduced as perturbations. Inspired by UKAN \cite{li2024u}, KANs and CNNs are integrated into both the encoder and decoder components to capture high-level semantic features as well as local image features.

\subsubsection{Convolution Block}
We apply three convolution blocks to feature maps characterized by weak high-resolution semantic information but rich spatial details. Each convolution block employs a $3 \times 3$ kernel size convolution, followed by batch normalization and ReLU activation functions, and can be expressed as:
\begin{equation}
O_{\mathrm{Conv}} = \mathrm{ReLU}(\mathrm{BN}(\mathrm{Conv}(x))),
\label{eq1}
\end{equation}
where $O_{\mathrm{Conv}}$ and $x$ represent the output and input of the convolution block, respectively. Semi-KAN adopts a U-Net backbone, which utilizes MLP-like architectures and incorporates activation functions (e.g., ReLU) to activate parameter weights. The process can be described as the interleaving of transformations $W$ and activations $\sigma$:
\begin{equation}
Y = (W_{L-1} \cdot \sigma \cdot W_{L-2} \cdot \sigma \cdot \ldots \cdot W_{1} \cdot \sigma \cdot W_{0})X,
\label{eq2}
\end{equation}
where $X$ is the input image, $X \in \mathbb{R}^{c \cdot h \cdot w}$, $c$ represents the number of channels, and $h$ and $w$ denote the height and width of the input image, respectively.

\subsubsection{KAN-Conv Block}
To enhance weight sharing and the feature representation of semantic information, we employ KAN layers \cite{liu2024kan} on feature maps with strong low-resolution semantic information. As shown in Fig.~\ref{fig3}, the output of the convolution block is reshaped using Tokenization \cite{dosovitskiy2020image} into a sequence of flattened 2D patches, which are then embedded into a latent $D$-dimensional space via a trainable linear projection $E \in \mathbb{R}^{(P^{2} \cdot C) \cdot D}$. Tokenization transforms the various feature maps into image patches. 

The proposed KAN-Conv Block incorporates depthwise convolution (DwConv) \cite{cao2022conv} and KAN layers, and the outputs from these components are combined to form the final result. The process can be expressed as:

\begin{figure}[t]
\centering
\includegraphics[width=0.5\textwidth]{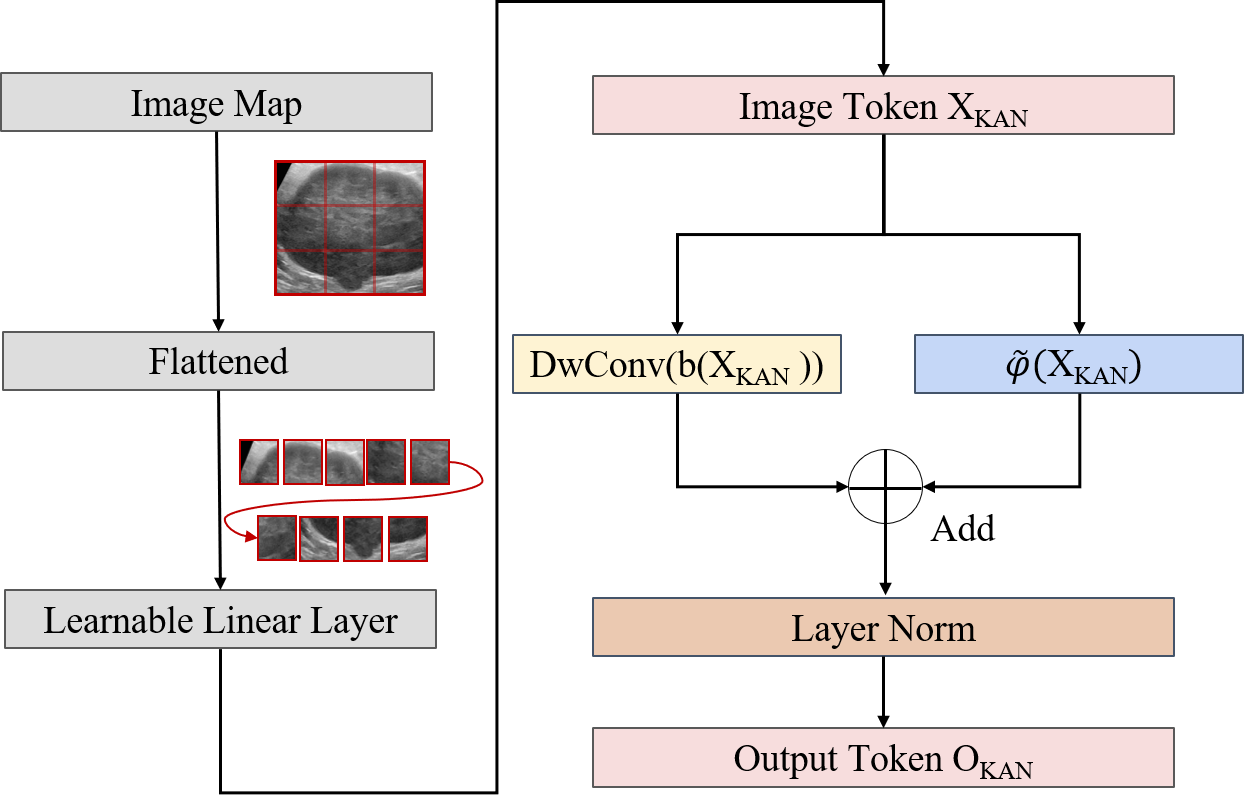} 
\caption{The architecture of KAN-Conv Block.}
\label{fig3}
\end{figure}

\begin{equation}
X^n_{\mathrm{KAN}} = E(\mathrm{Tokenization}(O_{\mathrm{Conv}})) \in \mathbb{R}^{(P^2 \cdot C) \cdot D},
\label{eq3}
\end{equation}

\begin{equation}
X_{\mathrm{KAN}} = \left[X^1_{\mathrm{KAN}}; X^2_{\mathrm{KAN}}; X^3_{\mathrm{KAN}}; \ldots; X^N_{\mathrm{KAN}}\right] \in \mathbb{R}^{N \cdot (P^2 \cdot C) \cdot D},
\label{eq4}
\end{equation}

\begin{equation}
Y_{n.i,j} = \sum_{n=1}^{N} \sum_{i=1}^{k-1} \sum_{j=1}^{k-1} \varphi_{n,j,i} X_{s+i,q+j}^n,
\label{eq5}
\end{equation}
where $P$ represents the patch size, and $N = hw / P^2$, with $h$ and $w$ denoting the height and width of the input. $k$ is the kernel size, $X^n_{s+i,s+j} \in X_{\mathrm{KAN}}$, $s = \overline{1,P^2C-k+1}$, and $q = \overline{1,D-k+1}$. Each $\varphi$ is a univariate non-linear learnable function with its own trainable parameters. When the input is $x$, the process is expressed as:

\begin{equation}
\varphi = w_b \cdot b(x) + \tilde{\varphi}(x),
\label{eq6}
\end{equation}

\begin{equation}
\tilde{\varphi}(x) = w_s \cdot \mathrm{Spline}(x),
\label{eq7}
\end{equation}

\begin{equation}
b(x) = \mathrm{SiLU}(x) = \frac{x}{1 + e^{-x}},
\label{eq8}
\end{equation}
where $w_b$ is a trainable parameter, $\mathrm{Spline}$ represents the spline function, and $\mathrm{SiLU}$ denotes the Sigmoid Gated Linear Unit function. In this work, we follow the original setting in \cite{liu2024kan}, which utilizes the spline function as the trainable activation function.

\subsection{Uncertainty Estimation-based Consistency Loss}
Semi-KAN employs multiple decoders, where different upsampling strategies are used as perturbations to enable robust representation learning. In this framework, we compute the variance across independent decoders and design an uncertainty estimation-based consistency loss to learn from unannotated data. Specifically, the mean of the outputs from different branches is used as a pseudo-annotation. The uncertainty estimation measures the similarity between different decoder outputs and the pseudo-annotation, which is then used to construct a consistency loss to guide the learning process.

\subsubsection{Uncertainty Evaluation}
Inspired by ensemble learning, we compute the mean of the outputs from the different decoders as a pseudo-annotation. Fuzzy classification probabilities tend to adversely affect performance, so we apply a sharpening function to adjust the output probability distribution, ensuring it lies in the high-confidence region. This process is described as:
\begin{equation}
p'(y|x; \varepsilon) = \frac{p(y|x; \varepsilon)^{1/T}}
{p(y|x; \varepsilon)^{1/T} + \left(1 - p(y|x; \varepsilon)\right)^{1/T}},
\label{eq9}
\end{equation}
\begin{equation}
\ p_{avg}^a=\frac{1}{B}\sum_{b=1}^{B}{p^\prime}_b^a. \
\label{eq10}
\end{equation}
where $x$ is the input to the segmentation model, $\varepsilon$ denotes the model parameters, and $p(y \mid x; \varepsilon)$ is the output of a decoder. $T$ is the hyperparameter controlling the temperature of the sharpening function, and $B$ represents the total number of decoders. ${p'}_b^a$ signifies the probability of pixel $a$ belonging to an organ in the output of decoder $b$. The similarity between decoder outputs and the pseudo-annotation is then calculated as:
\begin{equation}
\ U_b=\sum_{a=1}^{A}\sum_{b=1}^{B}{p^\prime}_b^a \cdot log\frac{{p^\prime}_b^a}{p_{avg}^a}, \
 \label{eq11}
\end{equation}

\subsubsection{Consistency Loss}
$U_b$ represents the uncertainty estimate of the output of decoder $b$. Based on this estimate, the consistency loss is defined as:
\begin{equation}
\ L_{consistency}=\partial L_{uncertainty}+(1-\partial)L_{rectify}, \
\label{eq14}
\end{equation}

\begin{equation}
\ L_{uncertainty}=\frac{1}{B}\sum_{b=1}^{B}U_b, \
\label{eq15}
\end{equation}

\begin{equation}
{{L}_{uncertainty}}=\frac{1}{B}\sum\limits_{b=1}^{B}{\frac{\sum\limits_{a=1}^{A}{{{\left\| {p^\prime}_{b}^{a}-p_{avg}^{a} \right\|}_{2}}\cdot \varpi _{b}^{a}}}{\sum\limits_{a=1}^{B}{\varpi _{b}^{a}}}}.
\label{eq16}
\end{equation}
where $\partial$ is a weighting factor balancing the contributions of $L_{\mathrm{uncertainty}}$ and $L_{\mathrm{rectify}}$. $\varpi_b^a$ is a rectifying weight, defined as $\varpi_b^a = e^{-U_b^a}$. The purpose of $L_{\mathrm{rectify}}$ is to emphasize reliable predictions while ignoring unreliable ones, ensuring stable self-supervised training. $L_{uncertainty}$ aims to reduce prediction entropy, further improving the robustness and generalization of the method.

\subsection{Interpretability of Semi-KAN}
KANs, as a form of symbolic representation, offer significant advantages in interpretability, which is a critical requirement in medical image analysis due to the need for reliable diagnostic outcomes. While recent studies have demonstrated the performance benefits of KANs in various domains \cite{drokin2024kolmogorov,vaca2024kolmogorov,tang20243d}, their potential for interpretability in image analysis tasks remains largely unexplored. Pruning methods proposed in the original work \cite{liu2024kan} face challenges in addressing the complexity and redundancy intrinsic to medical imaging data. To overcome these limitations, Semi-KAN introduces an interpretability-focused methods with two key contributions:  
\textbf{(1)} A shared encoder is employed to ensure consistency in high-level semantic features, while independent decoders are designed to capture distinct patterns through task-specific activation functions.  
\textbf{(2)} By utilizing a minimal number of feature channels and employing learnable B-spline-based activation functions, Semi-KAN facilitates the pruning of redundant nodes without compromising fidelity.

These architectural designs equip Semi-KAN with tools to enhance interpretability, thereby addressing a critical gap in medical image analysis tasks.

\begin{figure*}[t]
\centering
\includegraphics[width=1\textwidth]{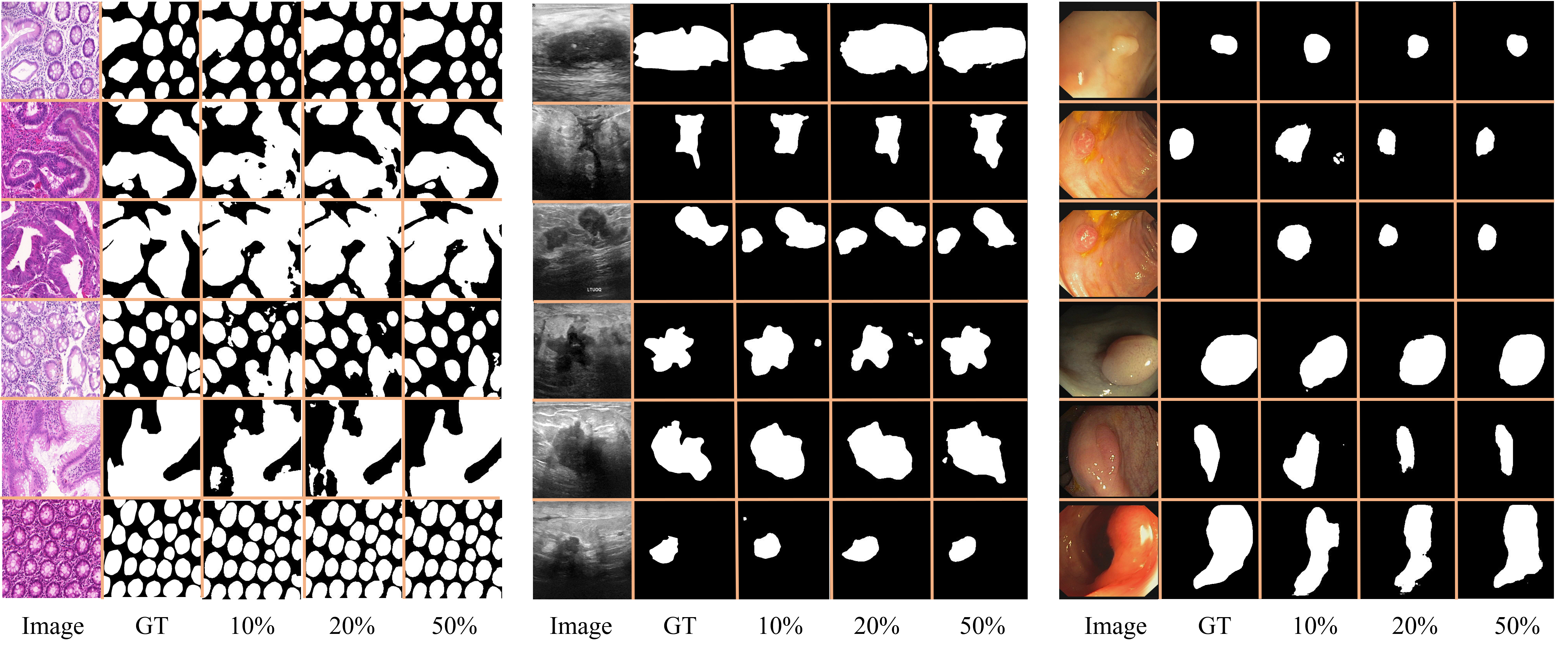} 
\caption{The segmentation results of Semi-KAN on the GlaS, BUSI and CVC datasets.}
\label{fig6}
\end{figure*}

\begin{table*}[htbp]
  \centering
  \caption{Performance Comparison on GlaS Dataset. Metrics with $\uparrow$ indicate higher is better, while $\downarrow$ indicates lower is better. Best results are in \textbf{bold}. The performance comparison values are calculated as the difference between Semi-KAN and the suboptimal or optimal algorithm.}
  \resizebox{\textwidth}{!}{%
  \begin{tabular}{llllll}
    \toprule
    \multicolumn{1}{l}{\multirow{2}{*}{\textbf{Method}}} & \textbf{\#Scans Used} & \multicolumn{4}{l}{\textbf{Metrics}} \\
    \cmidrule(lr){2-2} \cmidrule(lr){3-6}
    & $X_l$ / $X_u$ & Dice (\%) $\uparrow$ & Jaccard (\%) $\uparrow$ & HD95 (voxel) $\downarrow$ & ASD (voxel) $\downarrow$ \\
    \midrule
    Unet      & 100\% / 0\%   & \textcolor{gray}{91.65}         & \textcolor{gray}{84.93}          & \textcolor{gray}{2.02}          & \textcolor{gray}{1.05}          \\
    \midrule
    MC-net+   & 10\% / 90\%   & 80.91          & 70.14          & 6.60          & \textbf{3.78} \\
              & 20\% / 80\%   & 87.04          & 79.05          & 4.41          & 3.39          \\
              & 50\% / 50\%   & 88.31          & 81.15          & 2.96          & 1.28          \\
   
    AC-MT     & 10\% / 90\%   & 77.03          & 69.84          & 5.95          & 4.08          \\
              & 20\% / 80\%   & 81.92          & 73.43          & 4.90          & 3.77          \\
              & 50\% / 50\%   & 88.92          & 82.14          & \textbf{1.78}          & 1.04 \\
    
    Semi-KAN  & 10\% / 90\%   & \textbf{85.57} $\textcolor{green!70!black}{\uparrow} \scalebox{0.8}{\textcolor{green!70!black}{4.66}}$ & \textbf{74.79} $\textcolor{green!70!black}{\uparrow} \scalebox{0.8}{\textcolor{green!70!black}{4.65}}$ & \textbf{5.23} $\textcolor{green!70!black}{\downarrow} \scalebox{0.8}{\textcolor{green!70!black}{0.72}}$ & 3.89 $\textcolor{yellow!70!black}{\uparrow} \scalebox{0.8}{\textcolor{yellow!70!black}{0.11}}$ \\
              & 20\% / 80\%   & \textbf{90.31} $\textcolor{green!70!black}{\uparrow} \scalebox{0.8}{\textcolor{green!70!black}{3.27}}$ & \textbf{84.09} $\textcolor{green!70!black}{\uparrow} \scalebox{0.8}{\textcolor{green!70!black}{5.04}}$ & \textbf{2.64} $\textcolor{green!70!black}{\downarrow} \scalebox{0.8}{\textcolor{green!70!black}{1.77}}$ & \textbf{1.34} $\textcolor{green!70!black}{\downarrow} \scalebox{0.8}{\textcolor{green!70!black}{2.05}}$ \\
              & 50\% / 50\%   & \textbf{91.04} $\textcolor{green!70!black}{\uparrow} \scalebox{0.8}{\textcolor{green!70!black}{2.12}}$ &\textbf{84.51} $\textcolor{green!70!black}{\uparrow} \scalebox{0.8}{\textcolor{green!70!black}{3.97}}$ & \textbf{2.21} $\textcolor{yellow!70!black}{\uparrow} \scalebox{0.8}{\textcolor{yellow!70!black}{0.43}}$ & 1.35  $\textcolor{yellow!70!black}{\uparrow} \scalebox{0.8}{\textcolor{yellow!70!black}{0.30}}$ \\
    \bottomrule
  \end{tabular}%
  }
  \label{glas-results}
\end{table*}

\section{Experiments and Results}

\subsection{Datasets}
We evaluated the performance of Semi-KAN on four publicly available datasets: the Breast Ultrasound Segmentation Dataset (BUSI) \cite{al2020dataset}, the GlaS Challenge dataset \cite{valanarasu2021medical}, CVC-ClinicDB \cite{valanarasu2021medical}, and the Automated Cardiac Diagnosis Challenge (ACDC) dataset \cite{luo2021semi}. For BUSI, GlaS, and CVC-ClinicDB, we adhered to the preprocessing protocols outlined in \cite{li2024u}, while for ACDC, we followed the procedures described in \cite{luo2021semi}. All images were resampled to a resolution of $512 \times 512$ pixels. The experiments were conducted under annotation ratios of 10

\subsection{Implementation Details}
Semi-KAN was implemented using PyTorch and all experiments were conducted on a Linux-based system equipped with an NVIDIA GeForce RTX 3090 GPU (24GB memory). During training, we utilized the Adam optimizer with a batch size of 32, a momentum of 0.9, and a weight decay of 0.001. The model was trained for a maximum of 800 epochs.

\begin{table*}[htbp]
  \centering
  \caption{Performance Comparison on BUSI Dataset. Metrics with $\uparrow$ indicate higher is better, while $\downarrow$ indicates lower is better. Best results are in \textbf{bold}. The performance comparison values are calculated as the difference between Semi-KAN and the suboptimal or optimal algorithm.}
  \resizebox{\textwidth}{!}{%
  \begin{tabular}{lllllllll}
    \toprule
    \multicolumn{1}{l}{\multirow{2}{*}{\textbf{Method}}} & \textbf{\#Scans Used} & \multicolumn{4}{l}{\textbf{Metrics}} \\
    \cmidrule(lr){2-2} \cmidrule(lr){3-6}
    & $X_l$ / $X_u$ & Dice (\%) $\uparrow$ & Jaccard (\%) $\uparrow$ & HD95 (voxel) $\downarrow$ & ASD (voxel) $\downarrow$ \\
    \midrule
    Unet      & 100\% / 0\%   & \textcolor{gray}{77.59}         & \textcolor{gray}{63.38}          & \textcolor{gray}{3.86}         & \textcolor{gray}{1.81}          \\
    \midrule
    MC-net+   & 10\% / 90\%   & 60.96          & 46.22          & 12.11         & 5.37          \\
              & 20\% / 80\%   & 63.27          & 50.86          & 8.08          & 3.71          \\
              & 50\% / 50\%   & 70.32          & 54.31          & 7.83          & 3.25          \\
    
    AC-MT     & 10\% / 90\%   & 53.75          & 41.99          & \textbf{6.92} & 3.07          \\
              & 20\% / 80\%   & 67.66          & 51.13          & 6.77          & 2.96          \\
              & 50\% / 50\%   & 75.04          & 59.05          & 6.38          & \textbf{1.56} \\
    
    Semi-KAN  & 10\% / 90\%   & \textbf{64.21} $\textcolor{green!70!black}{\uparrow} \scalebox{0.8}{\textcolor{green!70!black}{3.25}}$ & \textbf{48.32} $\textcolor{green!70!black}{\uparrow} \scalebox{0.8}{\textcolor{green!70!black}{2.10}}$ & 8.33 $\textcolor{yellow!70!black}{\uparrow} \scalebox{0.8}{\textcolor{yellow!70!black}{1.41}}$ & \textbf{2.95} $\textcolor{green!70!black}{\downarrow} \scalebox{0.8}{\textcolor{green!70!black}{0.12}}$ \\
              & 20\% / 80\%   & \textbf{69.17} $\textcolor{green!70!black}{\uparrow} \scalebox{0.8}{\textcolor{green!70!black}{1.51}}$ & \textbf{52.87} $\textcolor{green!70!black}{\uparrow} \scalebox{0.8}{\textcolor{green!70!black}{1.74}}$ & \textbf{6.73} $\textcolor{green!70!black}{\downarrow} \scalebox{0.8}{\textcolor{green!70!black}{0.04}}$ & \textbf{2.57} $\textcolor{green!70!black}{\downarrow} \scalebox{0.8}{\textcolor{green!70!black}{0.39}}$ \\
              & 50\% / 50\%   & \textbf{75.49} $\textcolor{green!70!black}{\uparrow} \scalebox{0.8}{\textcolor{green!70!black}{0.45}}$ & \textbf{60.51} $\textcolor{green!70!black}{\uparrow} \scalebox{0.8}{\textcolor{green!70!black}{1.46}}$ & \textbf{6.02} $\textcolor{green!70!black}{\downarrow} \scalebox{0.8}{\textcolor{green!70!black}{0.36}}$ & 2.05 $\textcolor{yellow!70!black}{\uparrow} \scalebox{0.8}{\textcolor{yellow!70!black}{0.49}}$ \\
    \bottomrule
  \end{tabular}%
  }
  \label{busi-results}
\end{table*}

\begin{figure*}[t]
\centering
\includegraphics[width=1\textwidth]{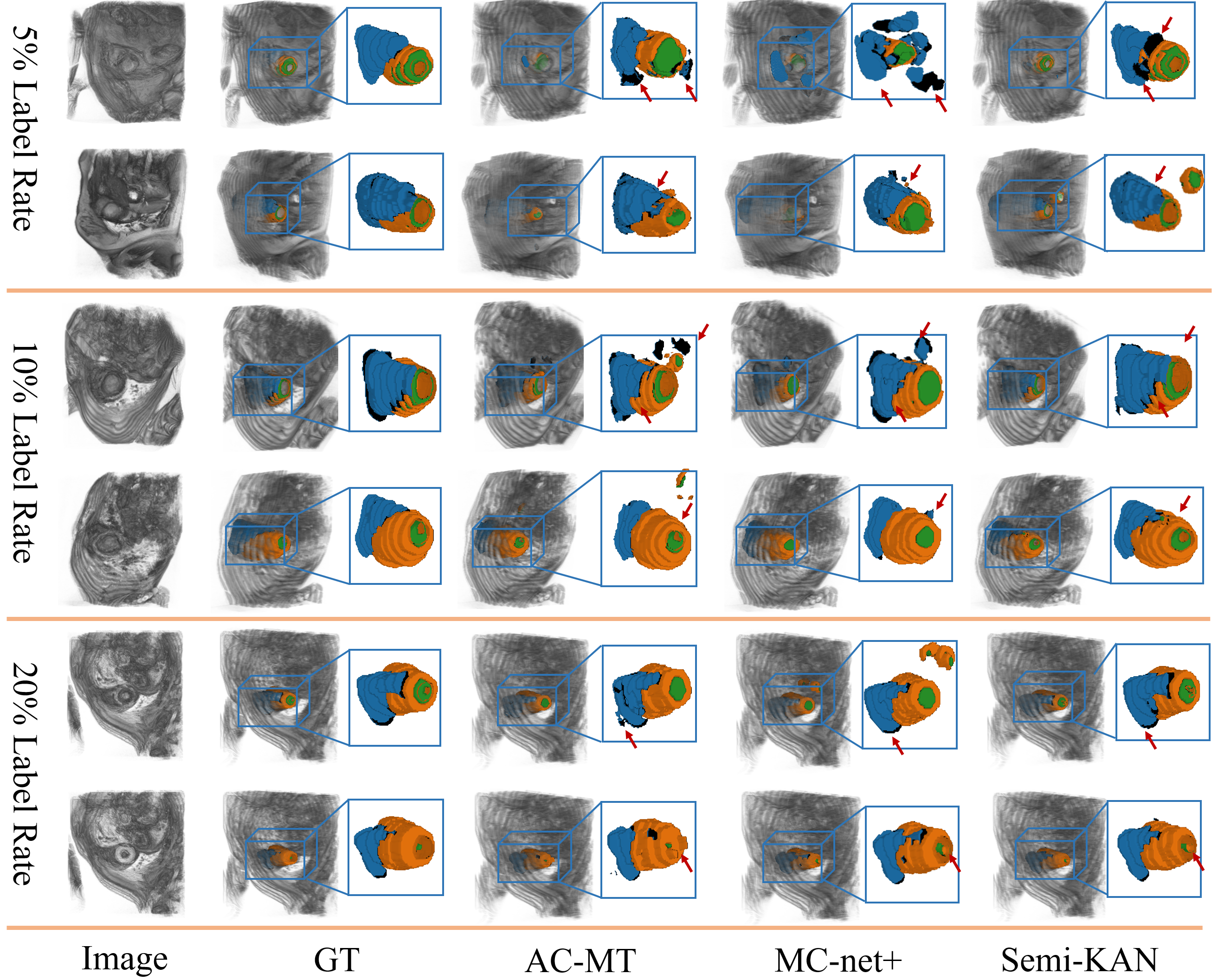} 
\caption{Qualitative comparisons of Semi-KAN, MC-net+ and AC-MT on the ACDC datasets.}
\label{fig7}
\end{figure*}

\begin{table*}[htbp]
  \centering
  \caption{Performance Comparison on CVC Dataset. Metrics with $\uparrow$ indicate higher is better, while $\downarrow$ indicates lower is better. Best results are in \textbf{bold}. The performance comparison values are calculated as the difference between Semi-KAN and the suboptimal or optimal algorithm.}
  \resizebox{\textwidth}{!}{%
  \begin{tabular}{lllllllll}
    \toprule
    \multicolumn{1}{l}{\multirow{2}{*}{\textbf{Method}}} & \textbf{\#Scans Used} & \multicolumn{4}{l}{\textbf{Metrics}} \\
    \cmidrule(lr){2-2} \cmidrule(lr){3-6}
    & $X_l$ / $X_u$ & Dice (\%) $\uparrow$ & Jaccard (\%) $\uparrow$ & HD95 (voxel) $\downarrow$ & ASD (voxel) $\downarrow$ \\
    \midrule
    Unet      & 100\% / 0\%   & \textcolor{gray}{89.18}          & \textcolor{gray}{83.79}          & \textcolor{gray}{2.69}          & \textcolor{gray}{0.97}          \\
    \midrule
    MC-net+   & 10\% / 90\%   & 51.26          & 38.56          & 11.24         & 5.25          \\
              & 20\% / 80\%   & 59.87          & 45.97          & 8.30          & 4.86          \\
              & 50\% / 50\%   & 69.83          & 55.94          & 7.36          & 4.07          \\
   
    AC-MT     & 10\% / 90\%   & 53.86          & 41.90          & 9.90          & 4.35          \\
              & 20\% / 80\%   & 76.49          & \textbf{69.98} & \textbf{5.03} & \textbf{3.40} \\
              & 50\% / 50\%   & 82.41          & 74.94 &\textbf{4.56} & \textbf{2.10} \\
   
    Semi-KAN  & 10\% / 90\%   & \textbf{54.41} $\textcolor{green!70!black}{\uparrow} \scalebox{0.8}{\textcolor{green!70!black}{0.55}}$ & \textbf{43.54} $\textcolor{green!70!black}{\uparrow} \scalebox{0.8}{\textcolor{green!70!black}{1.64}}$ & \textbf{8.78} $\textcolor{green!70!black}{\downarrow} \scalebox{0.8}{\textcolor{green!70!black}{1.12}}$ & \textbf{3.99} $\textcolor{green!70!black}{\downarrow} \scalebox{0.8}{\textcolor{green!70!black}{0.36}}$ \\
              & 20\% / 80\%   & \textbf{77.87} $\textcolor{green!70!black}{\uparrow} \scalebox{0.8}{\textcolor{green!70!black}{1.38}}$ & 68.97 $\textcolor{yellow!70!black}{\downarrow} \scalebox{0.8}{\textcolor{yellow!70!black}{1.01}}$ & 5.98 $\textcolor{yellow!70!black}{\uparrow} \scalebox{0.8}{\textcolor{yellow!70!black}{0.95}}$ & 3.64 $\textcolor{yellow!70!black}{\uparrow} \scalebox{0.8}{\textcolor{yellow!70!black}{0.24}}$ \\
              & 50\% / 50\%   & \textbf{82.96} $\textcolor{green!70!black}{\uparrow} \scalebox{0.8}{\textcolor{green!70!black}{0.55}}$ & \textbf{75.79} $\textcolor{green!70!black}{\uparrow} \scalebox{0.8}{\textcolor{green!70!black}{0.85}}$ & 5.76 $\textcolor{yellow!70!black}{\uparrow} \scalebox{0.8}{\textcolor{yellow!70!black}{1.20}}$ & 2.39 $\textcolor{yellow!70!black}{\uparrow} \scalebox{0.8}{\textcolor{yellow!70!black}{0.29}}$ \\
    \bottomrule
  \end{tabular}%
  }
  \label{cvc-results}
\end{table*}

\begin{table*}[htbp]
  \centering
  \caption{Performance Comparison on ACDC Dataset. Metrics with $\uparrow$ indicate higher is better, while $\downarrow$ indicates lower is better. Best results are in \textbf{bold}. The performance comparison values are calculated as the difference between Semi-KAN and the suboptimal or optimal algorithm.}
  \resizebox{\textwidth}{!}{%
  \begin{tabular}{lllllllll}
    \toprule
    \multicolumn{1}{l}{\multirow{2}{*}{\textbf{Method}}} & \textbf{\#Scans Used} & \multicolumn{4}{l}{\textbf{Metrics}} \\
    \cmidrule(lr){2-2} \cmidrule(lr){3-6}
    & $X_l$ / $X_u$ & Dice (\%) $\uparrow$ & Jaccard (\%) $\uparrow$ & HD95 (voxel) $\downarrow$ & ASD (voxel) $\downarrow$ \\
    \midrule
    Unet      & 100\% / 0\%   & \textcolor{gray}{91.65}           & \textcolor{gray}{84.93}           & \textcolor{gray}{1.89}          & \textcolor{gray}{0.56}          \\
    \midrule
    MC-net+   & 5\% / 95\%    & 60.77          & 50.19          & 15.98         & 5.07          \\
              & 10\% / 90\%   & 83.03          & 73.42          & \textbf{7.50} & 3.05          \\
              & 20\% / 80\%   & 85.51          & 74.76          & 5.09          & 1.92          \\
  
    AC-MT     & 5\% / 95\%    & 55.84          & 47.85          & 15.69         & 5.72          \\
              & 10\% / 90\%   & 81.68          & 71.40          & 9.06          & \textbf{2.28} \\
              & 20\% / 80\%   & 86.00          & 75.21          & \textbf{3.09} & \textbf{1.22} \\
   
    Semi-KAN  & 5\% / 95\%    & \textbf{65.57} $\textcolor{green!70!black}{\uparrow} \scalebox{0.8}{\textcolor{green!70!black}{4.80}}$ & \textbf{52.80} $\textcolor{green!70!black}{\uparrow} \scalebox{0.8}{\textcolor{green!70!black}{2.61}}$ & \textbf{14.67} $\textcolor{green!70!black}{\downarrow} \scalebox{0.8}{\textcolor{green!70!black}{1.02}}$ & \textbf{4.49} $\textcolor{green!70!black}{\uparrow} \scalebox{0.8}{\textcolor{green!70!black}{0.58}}$ \\
              & 10\% / 90\%   & \textbf{84.91} $\textcolor{green!70!black}{\uparrow} \scalebox{0.8}{\textcolor{green!70!black}{1.88}}$ & \textbf{75.04} $\textcolor{green!70!black}{\uparrow} \scalebox{0.8}{\textcolor{green!70!black}{1.62}}$ & 7.74 $\textcolor{yellow!70!black}{\uparrow} \scalebox{0.8}{\textcolor{yellow!70!black}{0.24}}$ & 2.32 $\textcolor{yellow!70!black}{\uparrow} \scalebox{0.8}{\textcolor{yellow!70!black}{0.04}}$ \\
              & 20\% / 80\%   & \textbf{88.95} $\textcolor{green!70!black}{\uparrow} \scalebox{0.8}{\textcolor{green!70!black}{2.95}}$ & \textbf{79.44} $\textcolor{green!70!black}{\uparrow} \scalebox{0.8}{\textcolor{green!70!black}{4.23}}$ & 4.67 $\textcolor{yellow!70!black}{\uparrow} \scalebox{0.8}{\textcolor{yellow!70!black}{1.58}}$ & 1.42 $\textcolor{yellow!70!black}{\uparrow} \scalebox{0.8}{\textcolor{yellow!70!black}{0.20}}$ \\
    \bottomrule
  \end{tabular}%
  }
  \label{acdc-results}
\end{table*}

\subsection{Main Results}

\subsubsection{Comparison with State-of-the-Art SSL Methods}
As shown in Tables \ref{glas-results}, \ref{busi-results}, \ref{cvc-results}, and \ref{acdc-results}, we benchmarked the performance of Semi-KAN against state-of-the-art semi-supervised learning (SSL) methods on the GlaS, BUSI, CVC, and ACDC datasets. Additionally, we reported the performance of U-Net trained with 100\% annotated samples, which serves as an upper bound. The evaluation metrics include Dice score, Jaccard, Hausdorff Distance 95th percentile (HD95), and Average Surface Distance (ASD). On the GlaS dataset, Semi-KAN achieves remarkable performance with 50\% annotated samples, obtaining a Dice score of 91.04\% and a Jaccard of 84.51\%, which is close to the upper bound performance achieved by U-Net with 100\% annotated samples. Furthermore, with only 10\% annotated samples, Semi-KAN achieves a Dice score of 85.57\% and a Jaccard of 74.79\%, representing improvements of 4.66\% and 4.65\%, respectively, compared to MC-Net+. On the BUSI dataset, segmentation proves to be significantly more challenging compared to GlaS, as all methods exhibit lower performance on this dataset. Semi-KAN maintains a performance advantage over other methods. With just 10\% annotated samples, Semi-KAN improves the Dice score from 60.96\% to 64.21\%, compared to MC-Net+. On the ACDC dataset, we further evaluated the performance of Semi-KAN with an even smaller fraction of annotated samples. Remarkably, with only 5\% annotated samples, Semi-KAN achieves a Dice score of 65.67\%, a Jaccard of 52.80\%, an HD95 of 14.67 voxels, and an ASD of 4.49 voxels. Compared to the second-best algorithm, these results represent improvements of 4.80\%, 2.61\%, 1.02 voxels, and 0.58 voxels, respectively.

\begin{table*}[htbp]
  \centering
  \caption{Performance Comparison of Different Activation Functions on GlaS Dataset. Metrics with $\uparrow$ indicate higher is better, while $\downarrow$ indicates lower is better. Best results are in \textbf{bold}.}
  \resizebox{\textwidth}{!}{%
  \begin{tabular}{lllllllll}
    \toprule
    \multicolumn{1}{c}{\multirow{2}{*}{\textbf{Method}}} & \textbf{\#Scans Used} & \multicolumn{4}{l}{\textbf{Metrics}} \\
    \cmidrule(lr){2-2} \cmidrule(lr){3-6}
    & $X_l$ / $X_u$ & Dice (\%) $\uparrow$ & Jaccard (\%) $\uparrow$ & HD95 (voxel) $\downarrow$ & ASD (voxel) $\downarrow$ \\
    \midrule
    Relu    & 10\% / 90\%   & 80.41          & 66.38          & 7.10          & 4.02
    \\
    & 20\% / 80\%   & 87.52          & 81.97          & 6.73          & 2.15          \\
                 & 50\% / 50\%   & 90.99          & 85.70          & 3.52          & 1.33          \\
    \midrule
    Att-Unet     & 10\% / 90\%   & 79.87          & 66.01          & 5.95          & 3.08          \\
    (Relu)       & 20\% / 80\%   & 85.43          & 76.75          & 3.30          & 2.32          \\
                 & 50\% / 50\%   & 88.01          & 83.18          & 2.55          & 1.58          \\
    \midrule
    Semi-KAN     & 10\% / 90\%   & \textbf{85.57}          & \textbf{74.79}          & \textbf{5.23}          & \textbf{3.89}          \\
    (KAN Layer)  & 20\% / 80\%   & \textbf{90.31} & \textbf{84.09} & \textbf{2.64} & \textbf{1.34} \\
                 & 50\% / 50\%   & \textbf{91.04} & \textbf{84.51} & \textbf{2.21} & \textbf{1.35} \\
    \bottomrule
  \end{tabular}%
  }
  \label{tab:activation-functions}
\end{table*}

\begin{figure*}[t]
\centering
\includegraphics[width=1\textwidth]{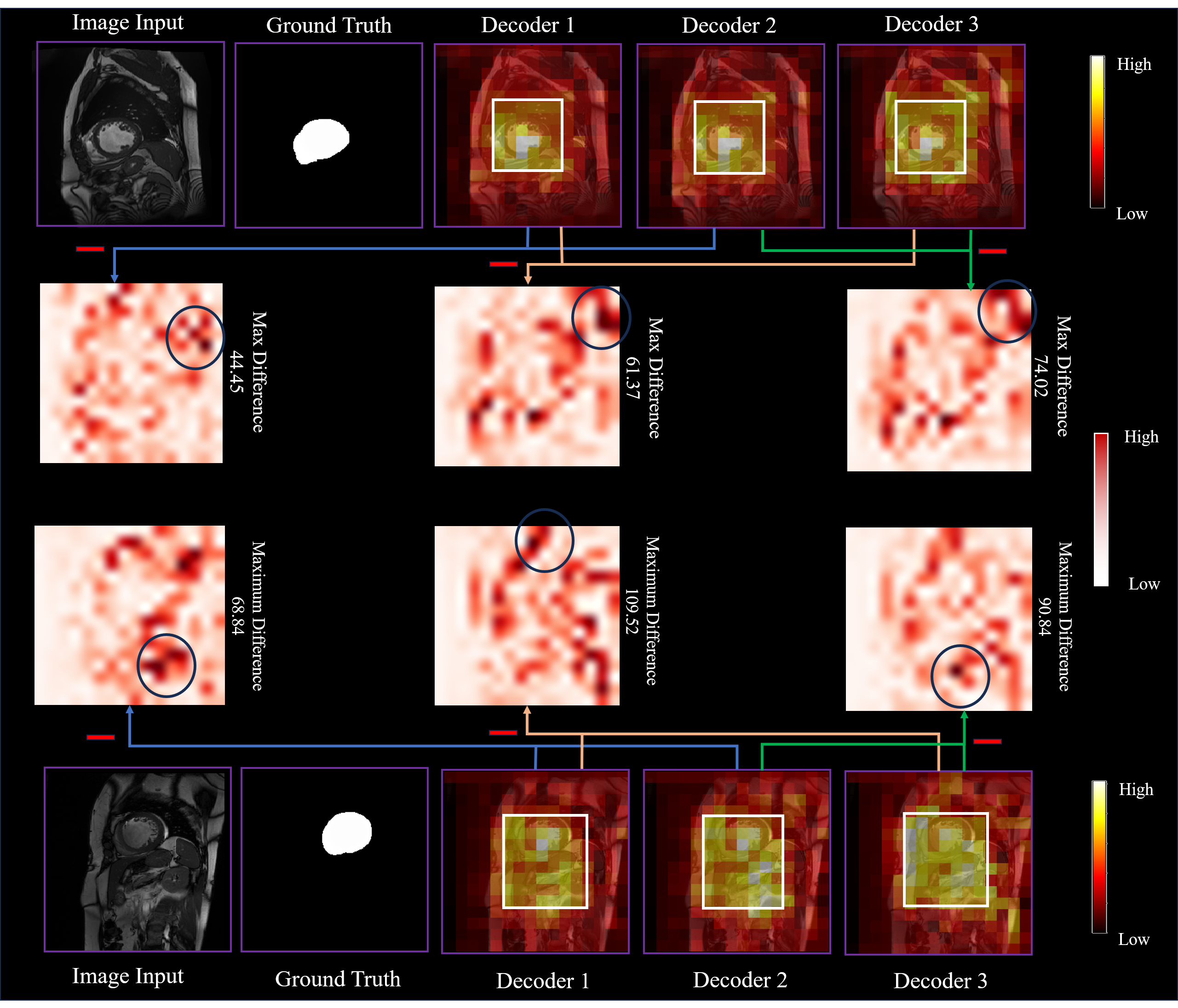} 
\caption{The highest semantic feature layer visualization of different decoders.}
\label{fig4}
\end{figure*}

\begin{figure*}[t]
\centering
\includegraphics[width=1\textwidth]{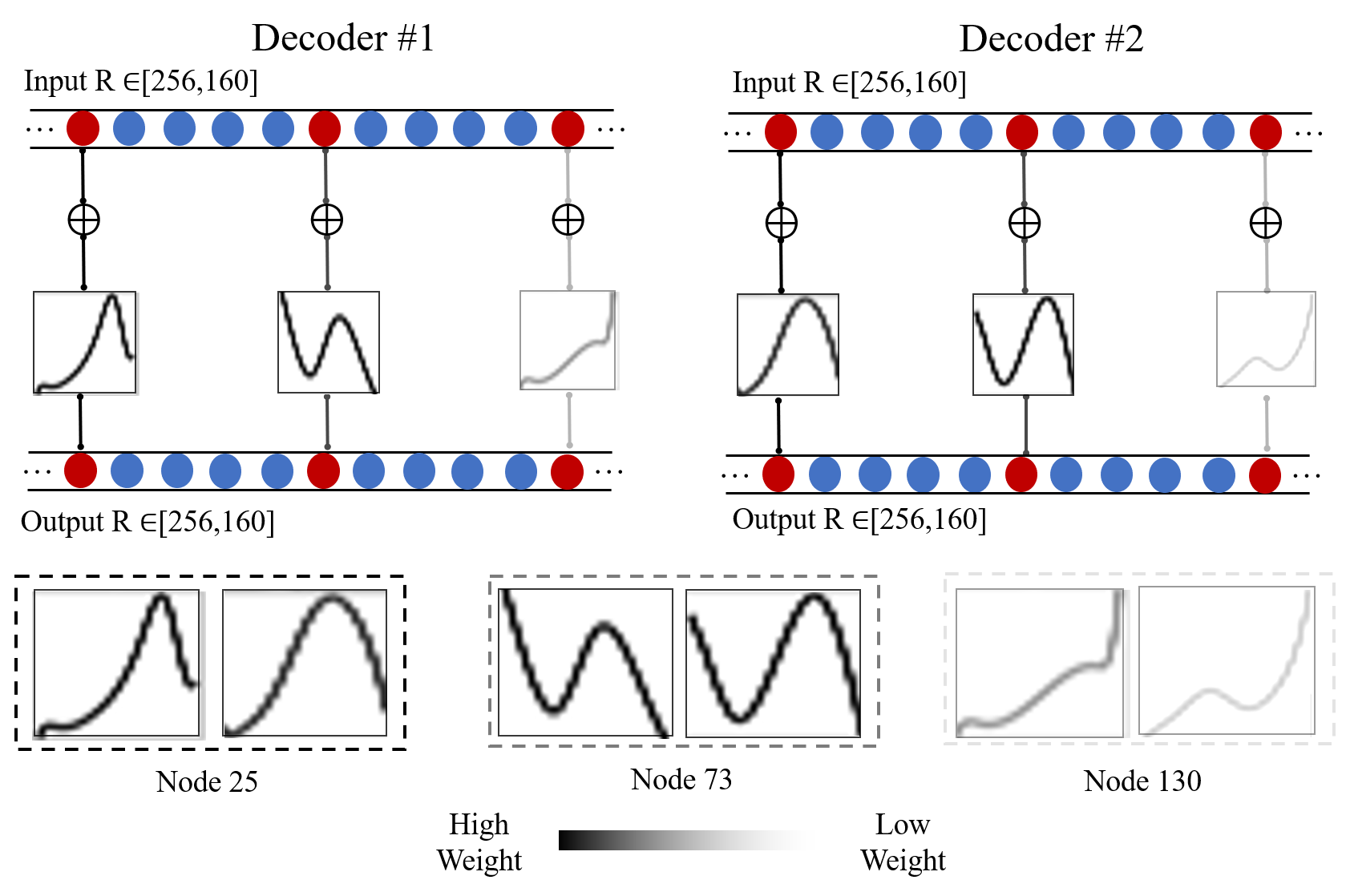} 
\caption{The activation functions visualization of different decoders.}
\label{fig5}
\end{figure*}

As shown in Fig.\ref{fig6}, we presented the segmentation results of Semi-KAN at varying labeling rates on the GlaS, BUSI and CVC datasets. Fig.\ref{fig7} depicts qualitative comparisons between Semi-KAN, MC-Net+, and AC-MT on the ACDC dataset. The visual results demonstrate that Semi-KAN effectively segments regions with varying sizes, shapes, and spatial locations.

These results indicate that Semi-KAN maintains strong segmentation accuracy even with very limited labeled data. Overall, Semi-KAN achieves significant segmentation performance across all four datasets, highlighting its excellent representational ability in semi-supervised learning scenarios.

Moreover, we observed that AC-MT exhibits a notable advantage in boundary segmentation. With only 10\% annotated samples, AC-MT achieves boundary segmentation accuracy comparable to that of Semi-KAN (Semi-KAN: 8.87 HD95, 3.99 ASD; AC-MT: 9.90 HD95, 4.35 ASD). However, as the labeling rate increases, AC-MT surpasses Semi-KAN in boundary segmentation accuracy. For instance, with 20\% annotated samples on the CVC dataset, AC-MT achieves improvements of 0.95 voxels in HD95 and 0.24 voxels in ASD over Semi-KAN. With 50\% annotated samples, these improvements further increase to 1.6 voxels in HD95 and 0.29 voxels in ASD. These improvements are attributed to AC-MT’s dual-network supervision model, which leverages Teacher and Student networks, as well as a specific boundary segmentation optimization strategy.However, this improvement in boundary accuracy comes at the cost of significantly increased computational demands. In contrast, Semi-KAN demonstrates better adaptability, achieving consistently strong segmentation performance across all four datasets. Notably, Semi-KAN excels in scenarios with limited labeled data, further underscoring its robustness and efficiency.

\subsubsection{Interpretability of Semi-KAN}
we visualized the highest semantic layer (spatial resolution: $16 \times 16$), upsampled it to $256 \times 256$, and superimposed the result onto the original image. This visualization reveals a strong alignment between the learned features and the anatomical boundaries. Further analysis of the decoder activation functions demonstrated consistent B-spline patterns across branches, indicating robust and reliable feature learning. These findings advance the interpretability of KANs in the context of medical imaging and underscore Semi-KAN's potential for trustworthy and clinically relevant segmentation tasks.

\subsection{Ablation Study}
\subsubsection{Effectiveness of Activation Functions}
We evaluated the impact of the learnable activation functions in Semi-KAN on the GlaS dataset. As shown in Table \ref{tab:activation-functions}, the learnable activation functions significantly enhance the performance of Semi-KAN compared to the rule-based (unlearnable) activation functions. Specifically, with a 10\% labeling rate on the GlaS dataset, the Dice score increased by 5.16\%, the Jaccard index by 8.41\%, and the HD95 decreased by 1.87 voxels. Similarly, at labeling rates of 20\% and 50\%, Semi-KAN continued to demonstrate a performance advantage. These results underscore the effectiveness of the learnable activation functions in improving segmentation accuracy and robustness.

\subsubsection{Effectiveness of Feature Representation Learning}
By incorporating KANs theory and a novel training pipeline, Semi-KAN achieves effective feature representation learning. As shown in Table \ref{tab:full-supervised-results}, we evaluated the feature representation capability of Semi-KAN by comparing its performance to U-KAN in a fully supervised setting on the ACDC, GlaS, BUSI, and CVC datasets. The results demonstrate that Semi-KAN achieves superior Dice and Jaccard compared to U-KAN on the GlaS, BUSI, and CVC datasets.

Additionally, we compared Semi-KAN with Att-Unet at different labeling rates on the GlaS dataset. The results reveal that Semi-KAN outperforms Att-Unet in terms of Dice score, Jaccard, and HD95, highlighting its robustness in semi-supervised learning scenarios. These findings confirm that Semi-KAN improves feature representation learning through its multi-decoder architecture and the introduction of an uncertainty estimation-based consistency loss.

\begin{table*}[htbp]
  \centering
  \caption{Performance Comparison on Different Datasets in Fully-supervised Learning. Metrics with $\uparrow$ indicate higher is better. Best results are highlighted in \textbf{bold}.}
  \resizebox{\textwidth}{!}{%
  \begin{tabular}{lcccccc}
    \toprule
    \multicolumn{1}{c}{\multirow{2}{*}{\textbf{Dataset}}} & \multicolumn{2}{l}{\textbf{\#Scans Used}} & \multicolumn{2}{c}{\textbf{U-KAN}} & \multicolumn{2}{c}{\textbf{Semi-KAN}} \\
    \cmidrule(lr){2-3} \cmidrule(lr){4-5} \cmidrule(lr){6-7}
    & $X_l$ & $X_u$ & Dice (\%) $\uparrow$ & Jaccard (\%) $\uparrow$ & Dice (\%) $\uparrow$ & Jaccard (\%) $\uparrow$ \\
    \midrule
    ACDC   & 100\% & 0\% & 91.65          &  \textbf{84.93}         & \textbf{92.05} $\textcolor{green!70!black}{\uparrow} \scalebox{0.8}{\textcolor{green!70!black}{0.40}}$ & 83.24 $\textcolor{yellow!70!black}{\downarrow} \scalebox{0.8}{\textcolor{yellow!70!black}{1.69}}$ \\
    GlaS   & 100\% & 0\% & 91.85          & 87.64          & \textbf{93.71} $\textcolor{green!70!black}{\uparrow} \scalebox{0.8}{\textcolor{green!70!black}{1.86}}$ & \textbf{88.64} $\textcolor{green!70!black}{\uparrow} \scalebox{0.8}{\textcolor{green!70!black}{1.00}}$ \\
    BUSI   & 100\% & 0\% & 75.73          & 63.38          & \textbf{79.02} $\textcolor{green!70!black}{\uparrow} \scalebox{0.8}{\textcolor{green!70!black}{3.29}}$ & \textbf{65.89} $\textcolor{green!70!black}{\uparrow} \scalebox{0.8}{\textcolor{green!70!black}{2.51}}$ \\
    CVC    & 100\% & 0\% & 88.92          & 85.05          & \textbf{90.92} $\textcolor{green!70!black}{\uparrow} \scalebox{0.8}{\textcolor{green!70!black}{2.00}}$ & \textbf{86.15} $\textcolor{green!70!black}{\uparrow} \scalebox{0.8}{\textcolor{green!70!black}{1.10}}$ \\
    \bottomrule
  \end{tabular}%
  }
  \label{tab:full-supervised-results}
\end{table*}

\section{Conclusions}
Kolmogorov-Arnold Networks (KANs) represent a transformative paradigm in neural network learning by leveraging stacks of non-linear, learnable activation functions. This unique design enhances feature learning capabilities while improving interpretability. In this study, we propose Semi-KAN, the first architecture integrating KANs into Semi-supervised Medical Image Segmentation (SSMIS). Semi-KAN adopts a multi-mode learning strategy, wherein convolutional blocks extract local features, while KANs are strategically applied at the encoder's bottleneck and the decoder's top layers to capture high-level semantic features. Extensive experiments conducted on four public datasets demonstrate the model's superior representation learning capabilities. Additionally, although previous studies have highlighted the interpretability advantages of KANs, specific methodologies for achieving interpretability have often been lacking. To address this gap, we propose a novel approach to enhance interpretability within the Semi-KAN framework. Overall, Semi-KAN underscores the potential of KANs in advancing SSMIS and promoting broader applications in medical image analysis.

%
%
%
\bibliographystyle{splncs04}
\bibliography{ourbib1}
%

\end{document}